\documentclass[letterpaper, 10 pt, conference]{ieeeconf}  

\IEEEoverridecommandlockouts                              

\usepackage[hidelinks, colorlinks=true, linkcolor=black, citecolor=green]{hyperref}
\usepackage{graphics} 
\usepackage{epsfig} 
\usepackage{times} 
\usepackage{lipsum}
\usepackage{amsmath} 
\usepackage{amssymb}  
\usepackage{verbatim} 
\usepackage{pifont}
\usepackage{blindtext}
\usepackage{xcolor} 
\usepackage{cuted}      
\usepackage{capt-of}    
\usepackage{graphicx}   
\usepackage{multirow}
\usepackage{booktabs}
\usepackage{url}
\usepackage{algorithm}
\usepackage{ragged2e}
\usepackage{cite}
\usepackage{algpseudocode}
\usepackage{booktabs,adjustbox}

\algnewcommand\Param{\item[\textbf{Parameter:}]}

\DeclareMathSizes{10}{9}{6}{5}

\title{\LARGE \bf
Beyond Object Selection: \\
Markerless Gaze-based Robot Placement at Arbitrary Positions

}

    \author{Yuzhi Lai$^{1*}$,~William Marx$^{4*}$, ~Shenghai Yuan$^{2}$,~Peizheng Li$^{1,3}$,~Zhuoyu Ran$^{1}$, and Andreas Zell$^{1}$
\thanks{Corresponding Author: \textbf{Andreas Zell}. $*$: Equal contribution.}
\thanks{$^{1}$University of Tuebingen,  Geschwister-Scholl-Platz, 72074 Germany, 
        {\tt\small \{name.surname\}@uni-tuebingen.de}.}%
\thanks{$^{2}$Nanyang Technological University, 50 Nanyang Avenue, Singapore 639798, 
        {\tt\small shyuan@ntu.edu.sg}.}%
        \thanks{$^{3}$
        {\tt\small peizheng.li@mercedes-benz.com}.}%
        \thanks{$^{4}$
        {\tt\small\{name.surname\}@student.uni-tuebingen.de}.}
}

\begin{document}

\bstctlcite{IEEEexample:BSTcontrol}

\maketitle
\thispagestyle{empty}
\pagestyle{empty}

\begin{abstract}

Gaze-based assistive manipulation typically supports object selection, while arbitrary-position placement requires
accurate spatial alignment between the headset and robot. However, for
gaze-based manipulation, pose accuracy does not necessarily translate into
task accuracy: translational and rotational errors jointly affect the
transformed gaze ray and may compensate for each other. To study
cross-device alignment from this task-oriented perspective, we present a
markerless interaction framework and a dedicated cross-device dataset. We
propose Graph-based Reference Selection to address sparse robot references. We further develop and benchmark multiple
task-specific alignment pipelines. Specifically, we introduce Gaze--Surface
Intersection Error (GSIE), which directly measures the spatial error of the
gaze-specified target. Experiments show that alignment methods ranked highly
by pose metrics are not always optimal in GSIE, demonstrating
the importance of evaluating gaze-based manipulation at the task level.

\end{abstract}

\section{INTRODUCTION}

People with severe motor impairments often retain voluntary eye movements and
speech, while unable to reliably use their arms for physical interaction,
making gaze a natural and accessible modality for assistive robot control
\cite{11127558, lai2026fam}. In this setting, to achieve seamless interaction, gaze-based interaction should allow users
not only to select an object, but also to indicate freely where it should be
placed. Supporting such arbitrary-position interaction requires the robot to
reliably align its workspace with the user's egocentric view, so that a
gaze-specified target can be transferred from the headset into the robot
coordinate frame.


Existing gaze-based systems typically avoid complete headset--robot
alignment by restricting interaction to object-level selection or predefined
placement targets~\cite{shahid2025gear,c7,lai2026fam}. Under these settings, gaze only needs to identify a known object. Such designs simplify cross-device correspondence,
but cannot support placement at arbitrary positions or more general
spatial manipulation. Fiducial Marker-based solutions~\cite{9889538} can provide direct headset--robot alignment, but require persistent marker visibility and are therefore vulnerable to occlusion while also constraining user's range of motion.

Markerless arbitrary-position placement introduces not only an alignment
problem, but also an evaluation problem. Sparse robot reference views and
large cross-device viewpoint changes make reliable
feature matching difficult, while existing benchmarks rarely provide paired
robot--headset observations for studying this setting. More importantly,
accurate pose estimation does not imply accurate placement.
Translation and rotation errors jointly affect the transformed gaze ray and
may partially compensate for or amplify each other. Consequently,
pose metrics alone cannot fully characterize the task accuracy for gaze-based manipulation.


To study this problem from a task-oriented perspective, we present a
markerless interaction framework for gaze-based object placement at arbitrary
positions. A robot-mounted RGB-D camera sparsely observes the workspace from
multiple viewpoints and reconstructs a 3D scene, providing calibrated reference
views for cross-device alignment. To avoid exhaustive feature matching over
all references, we propose Graph-based Reference Selection, which combines
semantic and spatial cues to select a compatible robot view before geometric
pose estimation. With this framework, we construct a dedicated cross-device
dataset and adapt point-based, line-assisted, and object-level refinement
pipelines to systematically benchmark different alignment strategies.
Finally, we introduce Gaze--Surface Intersection Error (GSIE), a task-oriented
metric that directly measures how alignment errors propagate to the final
gaze-specified placement position. The main contributions are summarized as
follows:

\begin{figure}
    \centering
    \includegraphics[width=0.45\textwidth]{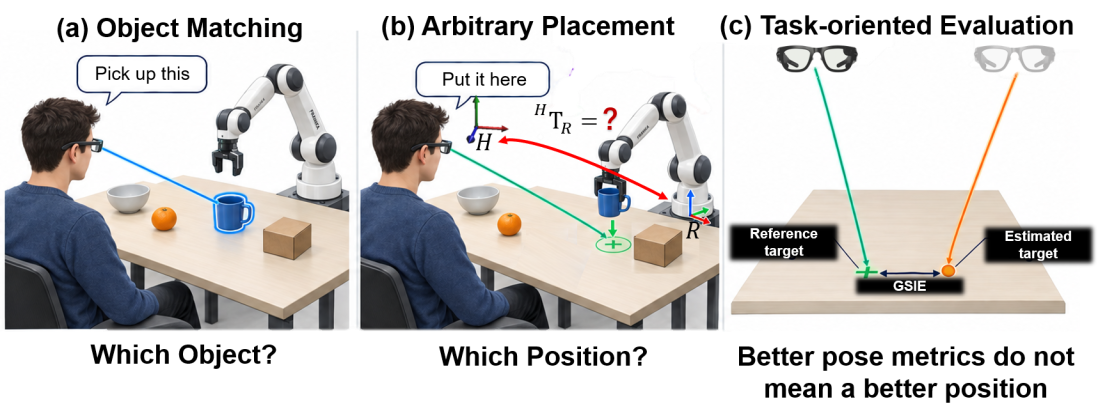}
    \vspace{-10pt}
    \caption{Concept art illustrating the motivation of our work: Robot placement at arbitrary positions and task-oriented evaluation.}
    \label{fig:1}
    \vspace{-10pt}
\end{figure}

\begin{itemize}
\item

We establish a markerless interaction framework and benchmark for
gaze-based placement at arbitrary positions, enabling systematic
evaluation of robot--headset alignment from pose- and task-level
perspectives.
\item We introduce a dedicated cross-device dataset with robot and headset
observations and propose Gaze--Surface Intersection Error (GSIE), a
task-oriented metric that measures the spatial deviation of the
gaze--surface intersection. Our dataset will be open sourced.
\item 

We propose a two-stage cross-device alignment approach that combines
graph-based reference selection with feature-based geometric pose estimation,
using semantic and spatial relationships to determine where cross-device
feature matching should be performed.

\end{itemize}

\section{Related Works}

Gaze provides an intuitive and low-effort interface for assistive robotic manipulation, allowing users to indicate targets directly through visual attention~\cite{11127558}. A key challenge in gaze-based robotic manipulation is aligning the user's
egocentric view with the robot workspace. Many existing systems \cite{c7, lai2026fam, shahid2025gear} avoid
explicit cross-device alignment by limiting gaze interaction to
object-level selection.
In \cite{shahid2025gear,tokmurziyev2025gazegrasp}, objects have distinct semantic labels, the robot can directly associate
the gazed object with its own detection. Feature-based approaches, such as
FAM-HRI~\cite{lai2026fam}, instead establish cross-view object correspondences
through local feature matching and nearest-neighbor assignment. However,
their reliability degrades under large viewpoint changes.
These approaches are effective for selecting \emph{what} to manipulate but
cannot directly specify \emph{where} an object should be placed when the
target is an arbitrary free position rather than an existing object or
predefined location. Fiducial markers such as ArUco or AprilTag can provide
the required headset--robot alignment~\cite{9889538,menendez2025semanticscanpath}, but require
persistent visibility and are unstable to occlusion. This motivates markerless cross-device alignment for arbitrary-position gaze-based placement.

Feature-based multi-perspective alignment has been widely studied in Simultaneous localization and mapping (SLAM) \cite{mur2015orb}, where maps are typically built from continuous trajectories with sufficient overlap between keyframes. 
For manipulator-mounted cameras, due to the robot geometry, limited workspace, and self-occlusion, continuously scanning the scene can introduce severe appearance changes and unstable tracking \cite{10128836}. This makes
appearance- or structure-based retrieval less reliable.
Graph-based approaches have also been explored for multi-perspective alignment.
In work\cite{lin2021topology,10128836, 9353207} authors construct graphs from
semantic and topological relations and directly estimate the relative pose
through graph matching. However, semantic and topological similarities can
produce incorrect correspondences, which may lead to significant pose errors.
GOReloc~\cite{wang2024goreloc} avoids directly estimating the pose from graph
matching, but constructs its matching using 2D query graph with a 3D object graph. Such
distances are not preserved under perspective projection. Moreover, its initial pose relies on
2D object centers and 3D object centroids, providing only coarse geometric
constraints; subsequent refinement may not fully compensate for errors
introduced by incorrect initialization.
Cross-device alignment is commonly evaluated using pose-based metrics, such as translational and rotational errors, or reprojection error~\cite{wang2024goreloc, xu2025airslam, voom}.  However, these metrics do not directly reflect the effect of pose estimation on the final
manipulation target. Translation and rotation errors can interact and
partially compensate for each other.

\section{Methodology}

\subsection{Problem Formulation}

The robot performs a horizontal circular scan of the workspace, recording a reference image sequence $\{\mathbf{I}^{C}_{i}\}_{i=1}^{N}$ and depth maps $\{\mathbf{D}^{C}_{i}\}_{i=1}^{N}$ with intrinsics ${}^{C}\mathbf{K}$. The headset records an egocentric sequence $\{\mathbf{I}^{H}_{j}\}_{j=1}^{M}$ with intrinsics ${}^{H}\mathbf{K}$, and the 2D gaze
targets ${}^{H}\mathbf{u}_j\in\mathbb{R}^{2}$ in the headset sequence. The camera extrinsic calibration ${}^{R}\mathbf{T}_{C}\in \mathrm{SE}(3)$ is known.
Given $\mathbf{I}^{H}_{j}$ and $\mathbf{I}^{C}_{i}$, the objective is
to estimate the transformation from the robot
base to the headset ${}^{H}\mathbf{T}_{R}$ and the 3D gaze
target in the robot base: ${}^R\mathbf{V}_j$.


\subsection{Interaction Framework}

\begin{figure}
    \centering
   
    \includegraphics[width=0.48\textwidth]{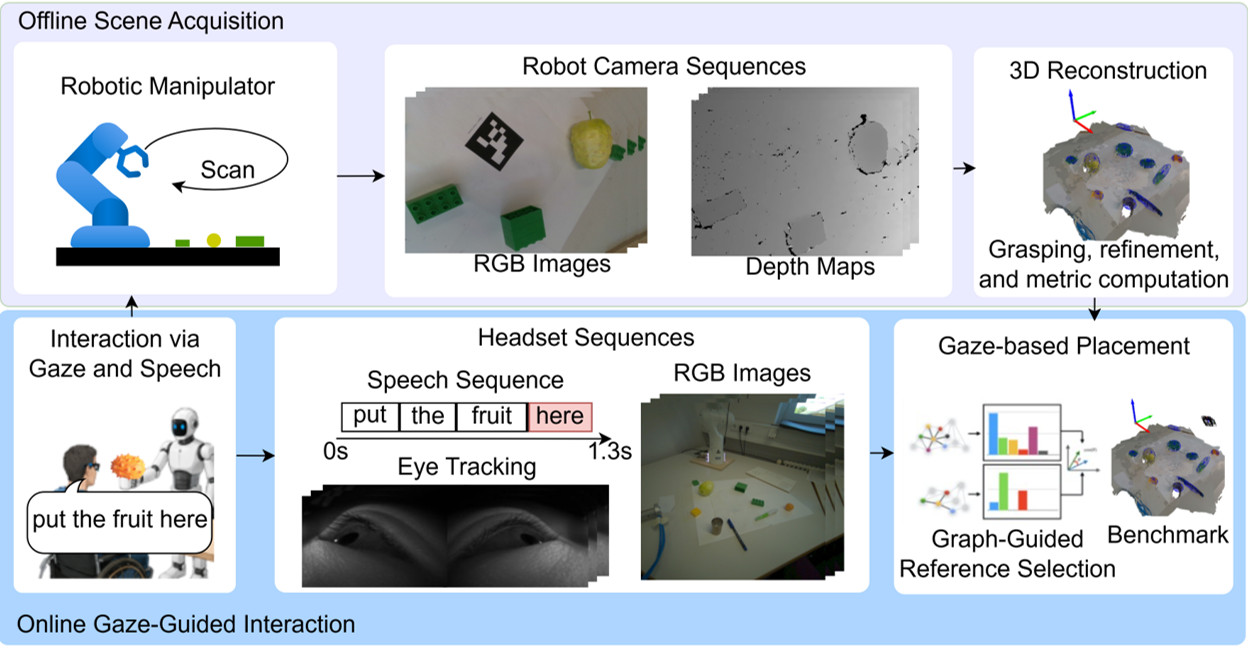}    
        \vspace{-15pt}
    \caption{Overview of the proposed benchmark framework. }
    \label{fig:pipeline}
\vspace{-15pt}
\end{figure}

As shown in Fig. \ref{fig:pipeline}, our framework consists of a scene-acquisition stage and a gaze-based interaction. During acquisition, we reconstruct the workspace from multiple RGB-D
observations by back-projecting $\mathbf{I}^{C}_{i}$ and
$\mathbf{D}^{C}_{i}$  into 3D via ${}^{R}\mathbf{T}_{C}^{-1}$. The resulting reconstruction is then aligned to the robot reference frame  using Kabsch–Umeyama alignment in $\mathrm{SIM}(3)$, followed by ICP to reduce inter-view layering artifacts and obtain the final fused 3D reconstruction $\mathcal{M}^{R} \in \mathbb{R}^{3\times n_i}$. By removing outliers via isolation forest\cite{10128836}, we obtain an object-level point cloud $\left\{ {}^{R}\mathbf{O}_i\in \mathbb{R}^{3\times n}\right\}_{i=0}^{o-1}$ ($o$ denotes the number of the objects) in robot base frame, which supports computation of task-level evaluation metrics. During interaction, the headset simultaneously records egocentric RGB images, eye-tracking signals, and the user's speech command. A deictic expression, such as ``here'', is identified from the command and associated with the placement action through dependency parsing \cite{hu2018vgpn, lai2026fam}. The gaze samples within the corresponding temporal interval are aggregated to obtain gaze points
${}^{H}\mathbf{u}_j$. 

\subsection{Graph-based Cross-Device Reference Selection}
\label{sec:graph}

Large viewpoint differences between the headset and robot cameras can produce unstable feature correspondences and inaccurate pose estimates. Existing object-graph methods \cite{wang2024goreloc} encode pairwise image distances, which vary considerably under perspective projection. We instead use coarse relative spatial relations (whether one object lies
to the left, right, above, or below another). This information is less sensitive to perspective changes and therefore provides a more stable cue for selecting a
compatible reference view.

As shown in Fig. \ref{fig:graph}, let $\mathcal{C}=\{c_i\}_{i=0}^{N_1-1}$ denote the set of object classes, where
$c_i^m$ represents the $m$-th detected instance of class $c_i$ with fixed sort. Each object forms a graph node, and every pair of nodes is connected by an edge
encoding their relative image-plane angle $\alpha$.
For each node $c_i^m$, we construct a semantic angle descriptor
$\mathrm{Des}(c_i^m)\in\mathbb{R}^{N_1\times N_2}$, where $N_2$ is the number of
angular bins $\mathcal{B}$. As shown in Fig. \ref{fig:graph}, $+1$ in $N_2$ ensures the consistency between the last bin and the first after full rotation. Its $(j,b)$-th entry counts the number of objects of class $c_j$
located within the $b$-th $\mathcal{B}$ relative to $c_i^m$:
\begin{figure}[t]
    \centering
    \includegraphics[width=0.45\textwidth]{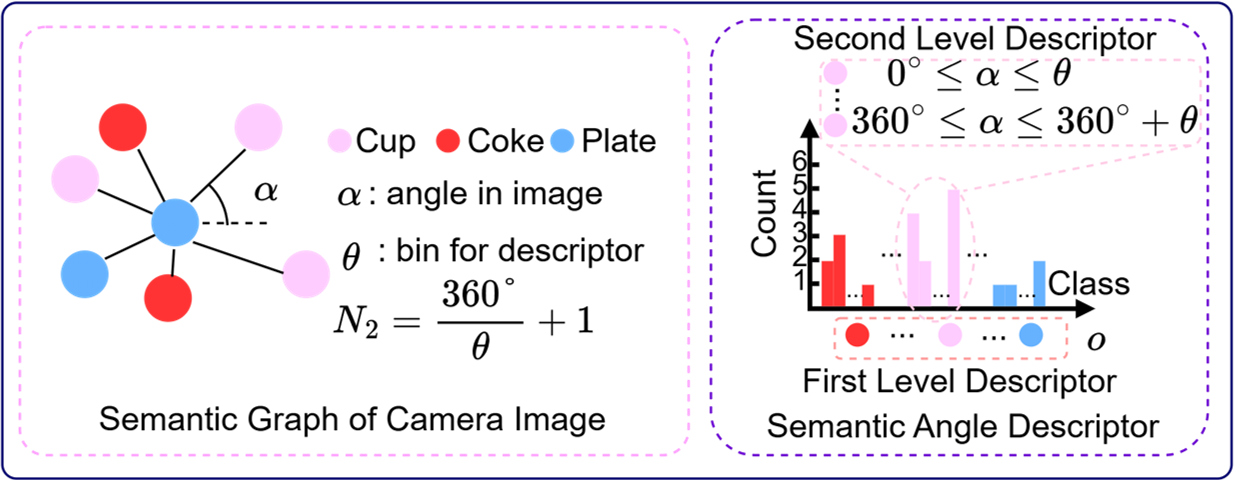}
    \caption{Overview of our graph-based cross-device reference selection. The first-level descriptor encodes semantic classes,
while the second-level descriptor captures the angular distribution of
semantic neighbors. 
}
    \label{fig:graph}
    \vspace{-15pt}
\end{figure}
\begin{equation}
\mathrm{Des}(c_i^m)_{j,b}
=
\sum_q\mathbb{I}
\left[
(j,q) \ne(i,m)
\right]
\mathbb{I}
\left[
\alpha(c_i^m,c_j^q)\in\mathcal{B}_b
\right]
\end{equation}
The descriptor of the $k$-th robot image graph $\mathrm{Des}_{R}^{k}(c_i^n)$ and the descriptor of the headset image graph $\mathrm{Des}_{H}(c_i^m)$ are
compared using matrix cosine similarity:
\begin{equation}
S_{imn}^{k}
=
\frac{
\left\langle
\mathrm{Des}_{H}(c_i^m),
\mathrm{Des}_{R}^{k}(c_i^n)
\right\rangle_F
}{
\left\|\mathrm{Des}_{H}(c_i^m)\right\|_F
\left\|\mathrm{Des}_{R}^{k}(c_i^n)\right\|_F
}
\end{equation}
Where $\left\langle\cdot,\cdot\right\rangle_F$ represents Frobenius inner product. For each headset image graph, its best match score with $k$-th robot image graph is a bipartite graph matching problem. We then compute the globally optimal one-to-one matching $\mathcal{X}_k$ with Hungarian Algorithm.
The compatibility score of robot image graph $k$ is computed with the number of detected object (via YOLO26 \cite{jocher2026ultralyticsyolo26unifiedrealtime}) in headset image $N$:

\begin{equation}
\bar{S}_k
=
\frac{1}{N}
\sum_{(i,m,n)\in\mathcal{X}_k}
S_{imn}^{k}.
\end{equation}
Finally, the robot image with the highest compatibility score is selected as
the reference image:

\begin{equation}
k^*
=
\arg\max_k \bar{S}_k.
\end{equation}
The selected reference provides the most compatible spatial
configuration for subsequent feature matching and pose estimation. Since the match between the robot reference image and ${}^{R}\mathbf{O}_i$ is deterministic, $\mathcal{X}_k$ also represents the match between the 2D object in the headset images and ${}^{R}\mathbf{O}_i$.

\subsection{Object-Level Refinement via Enclosing Ellipsoid}
\label{sec:ellipsoid_refinement}

Even if the best reference robot image is selected, Perspective-n-Point (PnP) may remain inaccurate under unstable feature
correspondences. 
Following established
quadric representations~\cite{10128836}, our formulation
adapts this geometry to cross-device alignment and gaze-guided placement. We adopt the quadric representation because both the projected 3D quadric
and the observed 2D ellipse admit equivalent Gaussian representations, providing a common probabilistic space.
With
$\left\{ {}^{R}\mathbf{O}_i\right\}_{i=0}^{o-1}$, each object is represented
by its minimum-volume enclosing ellipsoid (MVEE), obtained by:

\begin{equation}
\begin{aligned}
\min_{\mathbf{A}\succ 0,\,\mathbf{c}}
&\quad -\log\det(\mathbf{A}) \\
\mathrm{s.t.}
&\quad
({}^{R}\mathbf{p}-\mathbf{c})^{\top}
\mathbf{A}
({}^{R}\mathbf{p}-\mathbf{c})
\leq 1,
\quad
\forall\,{}^{R}\mathbf{p}\in{}^{R}\mathbf{O}_{i},
\end{aligned}
\end{equation}
where $\mathbf{c}$ is the ellipsoid center and the symmetric positive-definite matrix $\mathbf{A}$ determines
its orientation and semi-axis lengths. The resulting primal quadrics ${}^{R}\mathbf{Q}_i=\left [\begin{smallmatrix}
 \mathbf{A} & -\mathbf{A}\mathbf{c} \\
 -\mathbf{c}^{\top}\mathbf{A} &\mathbf{c}^{\top}\mathbf{A}\mathbf{c}-1
\end{smallmatrix}\right ] \in \mathbb{R}^{4\times 4}$ form an object map $\left\{ {}^{R}\mathbf{Q}_i\right\}_{i=0}^{o-1}$. 
MVEE directly optimizes a
tight enclosing volume via Khachiyan algorithm \cite{khachiyan1996rounding}.
For PCA-based fitting \cite{voom}, its
independently estimated principal-axis lengths do not guarantee that all
observed points are enclosed. In contrast, MVEE explicitly enforces
point-wise containment while minimizing the enclosing volume.
\begin{figure}
    \centering
    \includegraphics[width=0.45\textwidth]{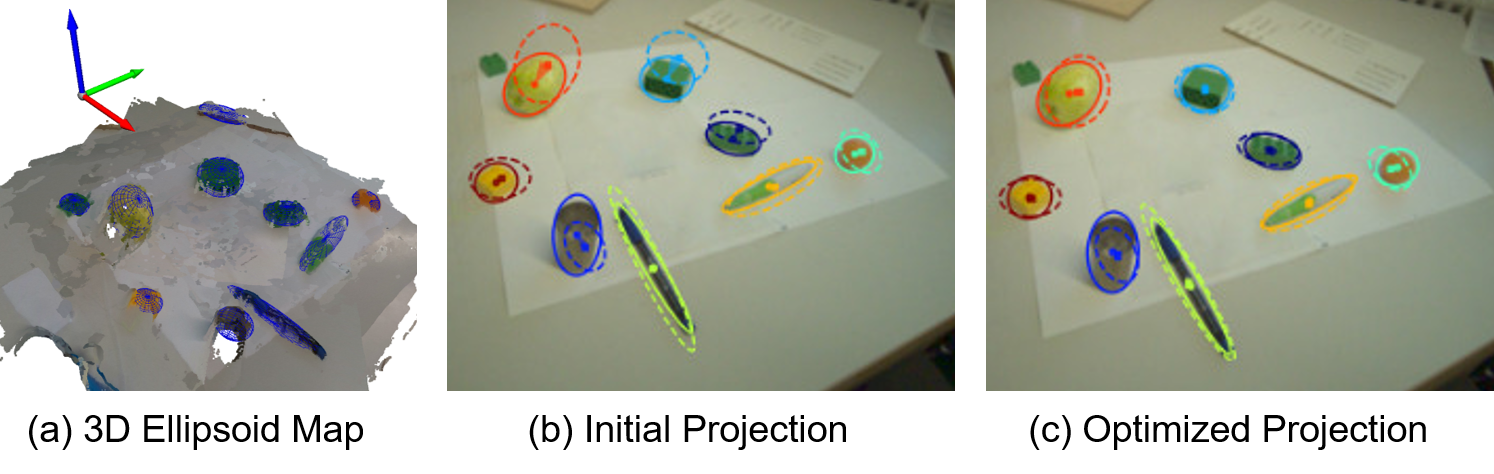}
    \vspace{-10pt}
    \caption{Object-level ellipsoid refinement before and after pose optimization. 
}
    \label{fig:ellipsiod}
    \vspace{-5pt}
\end{figure}
For a headset query image, object masks are fitted with 2D ellipses ${}^{H}E_i$. Given the initial PnP estimate ${}^{H}\mathbf{T}_R=\left [\begin{smallmatrix}
 {}^{H}\mathbf{R}_R & {}^{H}\mathbf{t}_R \\
 0 &1
\end{smallmatrix} \right ]  \in \mathbb{R}^{4\times 4} $, the projection matrix is defined as ${}^{H}\mathbf{P}_R
=
{}^{H}\mathbf{K}
[\begin{smallmatrix}
{}^{H}\mathbf{R}_{R} &
{}^{H}\mathbf{t}_{R}
\end{smallmatrix}]$.
A primal quadric ${}^{R}\mathbf{Q}$ is projected onto the headset image as

\begin{equation}
{}^{H}\mathbf{E}_{\mathrm{proj}}
=
\left(
{}^{H}\mathbf{P}_R\,
{}^{R}\mathbf{Q}_i^{-1}
{}^{H}\mathbf{P}_R^{\top}
\right)^{-1}
\end{equation}
Each primal conic
$\mathbf{E}=
\left [\begin{smallmatrix}
\mathbf{A} & \mathbf{b}\\
\mathbf{b}^{\top} & c
\end{smallmatrix}\right ]$
is represented by a Gaussian $\mathcal{N}(\boldsymbol{\mu},\boldsymbol{\Sigma})$, where
$
\boldsymbol{\mu}
=
-\mathbf{A}^{-1}\mathbf{b},
\boldsymbol{\Sigma}
=
\left(
\frac{\mathbf{A}}
{\boldsymbol{\mu}^{\top}\mathbf{A}\boldsymbol{\mu}-c}
\right)^{-1}.
$
As illustrated in Fig.~\ref{fig:ellipsiod}, projected 3D quadrics and the detected ellipses of the matched objects $\mathcal{X}_k$ (from Sec. \ref{sec:graph}) are matched by minimizing
their squared Wasserstein distance $\mathcal{W}=W_{2}^{2}(\mathcal{N}({}^{H}\mathbf{E}_{\mathrm{proj}}),\mathcal{N}({}^{H}\mathbf{E}_{i}))$.
The
pose is refined by:
\begin{equation}
\widehat{{}^{H}\mathbf{T}_{R}}
=
\arg\min_{{}^{H}\mathbf{T}_{R}}
\sum_{({}^{R}\mathbf{Q}_{i},\,{}^{H}\mathbf{E}_{i})
\in\mathcal{X}_{k}}
\mathcal{W}
\end{equation}
With multiple object correspondences, a small number of imperfect masks has
limited influence because the remaining objects still provide consistent
constraints.


\subsection{Perspective-n-Point-Line}
\begin{figure}
    \centering
    \includegraphics[width=0.45\textwidth]{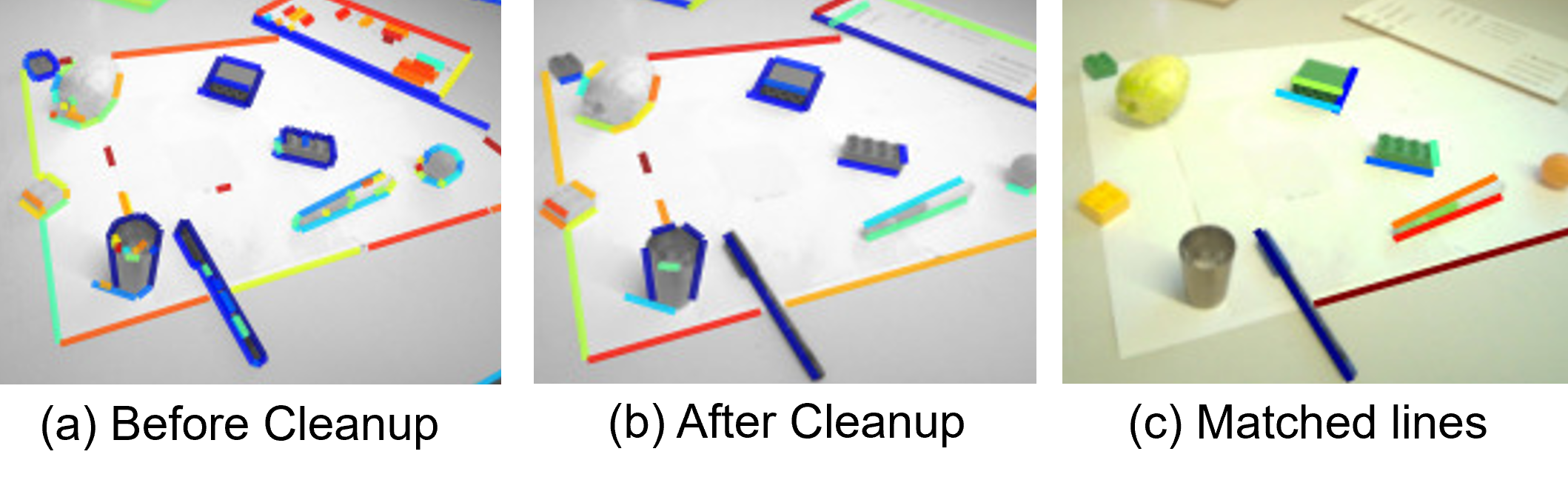}
    \vspace{-10pt}
    \caption{PnP+Line Alignment Pipeline. Structural lines are extracted and matched
across the robot and headset views.}
    \label{pnpl}
    \vspace{-15pt}
\end{figure}
To improve pose estimation in scenes containing weakly textured but structurally salient regions,
we extend the initial PnP estimate with 3D--2D line correspondences~\cite{xu2023airvo}.


\begin{table}[t]
    \centering
    \vspace{7pt}
    \refstepcounter{table}
    \label{tab:dataset_scenes}

    \begin{minipage}{\columnwidth}
        \footnotesize
        \centering
        \textbf{Table~\thetable.}
        Overview of our cross-device dataset.
    \end{minipage}
    \par\vspace{0.8ex}

    \normalsize
    \setlength{\tabcolsep}{2.6pt}
    \renewcommand{\arraystretch}{1.08}

    \begin{adjustbox}{max width=\columnwidth}
        \begin{tabular}{@{}cllc@{}}
            \toprule
            \textbf{Scene}
            & \textbf{Geometry}
            & \textbf{Object types}
            & \textbf{Count} \\
            \midrule

            $S_A$
            & Rounded / ellipsoidal
            & Fruits, pens, dishes, bowls
            & 5--12 \\

            $S_B$
            & Cuboid / box-shaped
            & LEGO Duplo bricks, boxes
            & 10--17 \\

            $S_C$
            & Mixed / irregular
            & Mixed everyday objects
            & 6--13 \\
            \bottomrule
        \end{tabular}
    \end{adjustbox}

    \par\vspace{-18pt}
\end{table}

As shown in Fig. \ref{pnpl}, for each robot reference image, line segments are extracted using the Line Segment Detector (LSD)~\cite{von2008lsd}. To make the geometric thresholds
consistent across cameras with different resolutions, all pixel coordinates
are normalized by the image diagonal. Short and fragmented segments are removed or merged according to their angular, midpoint, and endpoint distances \cite{xu2023airvo}. Using the aligned depth map and the 3D reconstruction, the pixels associated with each retained 2D segment $\ell$ are back-projected into 3D. The depth pixels associated with a 2D line often contain outliers due to
foreground--background discontinuities, missing depth values, and sensor
noise. We therefore use RANSAC to robustly fit a 3D line $\mathbf{L}(t)=\mathbf{a}+t\mathbf{b}$ from the
back-projected points $\mathcal{P}_{\ell}$, retaining only points consistent with the dominant
linear structure:

\begin{equation}
(\mathbf{a}^{*},\mathbf{b}^{*})
=
\arg\max_{\mathbf{a},\,\|\mathbf{b}\|_2=1}
\sum_{\mathbf{p}\in\mathcal{P}_{\ell}}
\mathbb{I}
\left[
\left\|
(\mathbf{p}-\mathbf{a})\times\mathbf{b}
\right\|_2
<\tau_{\mathrm{3D}}
\right]
\label{8}
\end{equation}
Where $\mathbf{a}$ is a point on the candidate line, $\mathbf{b}$ is its
unit direction vector, and $\tau_{\mathrm{3D}}$ is the inlier threshold.
Given $\mathbf{I}^{H}$, the selected robot reference image  $\mathbf{I}_k^{C}$ (from Sec. \ref{sec:graph}), 3D reconstruction, and the 2D-2D point matching, we get the 2D-3D matching $\mathcal{X}_{P}=
\left\{ \left(
{}^{R}\mathbf{p}_{q},
{}^{H}\mathbf{u}_{q},{}^{C}\mathbf{u}_{q}
\right) \right\}_{q=0}^{N_p-1}$. ${}^{H}\mathbf{u}_{q}$ and ${}^{C}\mathbf{u}_{q}$ represent the 2D point in headset image and robot camera image, respectively. ${}^{R}\mathbf{p}_{q}$ represents the 3D point in robot base frame. $N_p$ denotes the number of the matched features.
Let
${}^{C}\ell_i$ and ${}^{H}\ell_j$ denote the 2D lines in the robot and headset
image, respectively. Their similarity is defined as the number of matched
point pairs lying close to both lines:
\begin{equation}
        s_{ij}
=
\sum_{q=0}^{N_q-1}
\mathbb{I}
\left[
d({}^{C}\mathbf{u}_{q},{}^{C}\ell_i)<\Delta
\right]
\mathbb{I}
\left[
d({}^{H}\mathbf{u}_{q},{}^{H}\ell_j)<\Delta
\right] 
\label{9}
\end{equation}
Where $d(\cdot,\cdot)$ denotes the distance function from a point to a line segment. $\Delta$ represents the distance threshold. A line pair is retained when its score exceeds a minimum threshold
$\tau_L$ and is the distinctive mutual-best match for both lines. The 3D--2D line correspondences $\mathcal{X}_{L}
=
\left\{
\left(
{}^{R}\mathbf{L}_{i},
{}^{H}\ell_{i}
\right)
\right\}_{i=0}^{N_{\mathbf{L}}-1}$ are derived from Eq. \ref{8} and \ref{9}, where ${}^{R}\mathbf{L}_{i}$ is the 3D line segment in robot base frame associated with
${}^{C}\ell_i$. $N_{\mathbf{L}}$ represents the number of the corresponding lines.
The initial PnP estimate is refined by jointly minimizing the point
reprojection error and the projected-line error:   

\begin{align}
    e_{\mathrm{p}}
&=
\sum_{({}^{R}\mathbf{p},{}^{H}\mathbf{u})\in\mathcal{X}_{P}}
\left\|
\pi\left(
{}^{H}\mathbf{T}_{R}{}^{R}\mathbf{p},
{}^{H}\mathbf{K}
\right)
-
{}^{H}\mathbf{u}
\right\|_2 \notag
\\
e_{\mathrm{l}}
&=
\sum_{({}^{R}\mathbf{L},{}^{H}\ell)\in\mathcal{X}_{L}}
\sum_{{}^{R}\mathbf{p}\in\operatorname{end}({}^{R}\mathbf{L})}
d\left(
\pi\left(
{}^{H}\mathbf{T}_{R}{}^{R}\mathbf{p},
{}^{H}\mathbf{K}
\right),
{}^{H}\ell
\right) \notag
\\
e
&=
\frac{1-\lambda}{|\mathcal{X}_{P}|}e_{\mathrm{p}}
+
\frac{\lambda}{|\mathcal{X}_{L}|}e_{\mathrm{l}}
\label{pnplerror}
\end{align}
Where $\pi(\cdot)$ denotes the reprojection function, $\operatorname{end}$ represents the endpoints, and $\lambda$ balances the point and line constraints. The pose ${}^{H}\mathbf{T}_R$ is
optimized via Levenberg--Marquardt.



\begin{figure}
    \centering
    \includegraphics[width=0.45\textwidth]{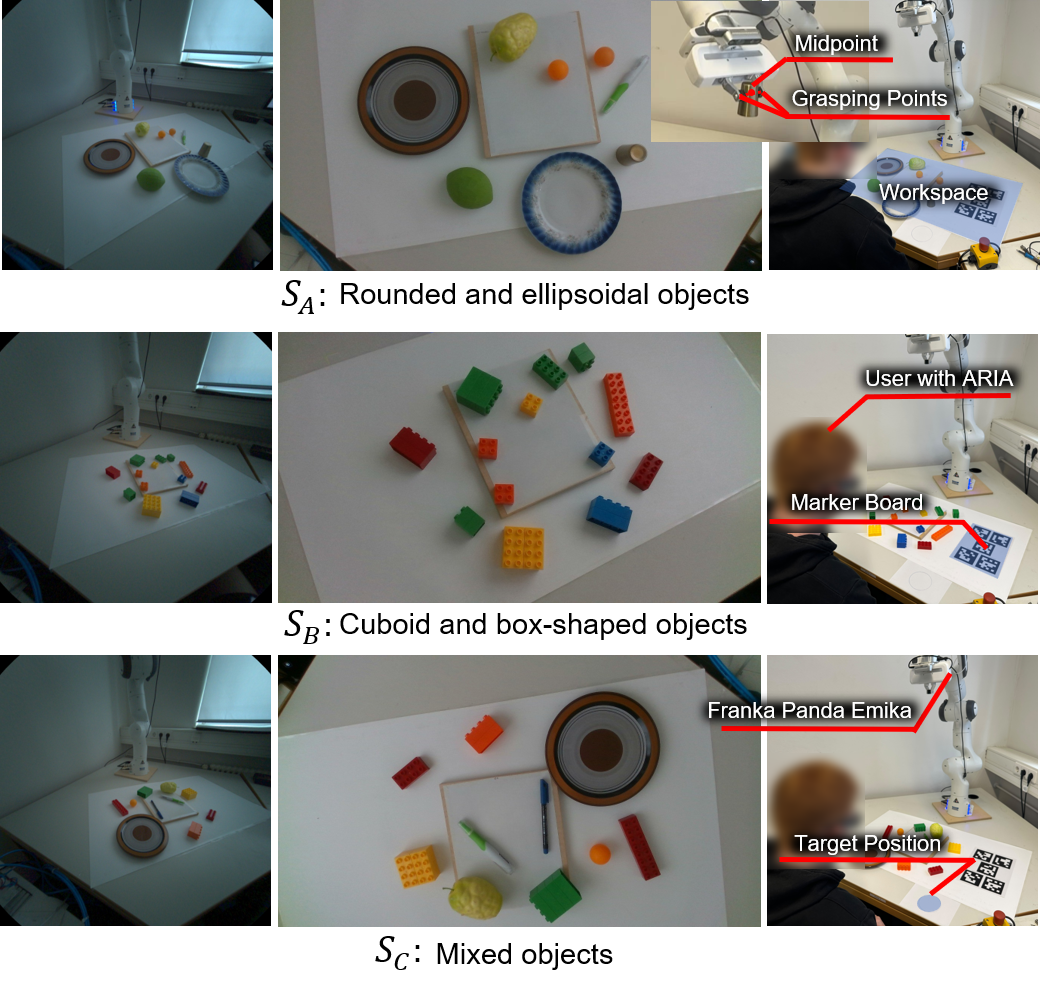}
    \vspace{-10pt}
    \caption{Cross-Device Dataset. Each scene contains 10 different sequences. Left: Glasses View. Mid: Robot View. Right: Third View}
    \label{data}
    \vspace{-15pt}
\end{figure}

\section{Cross-Device Dataset}
We construct a cross-device dataset using a Franka Emika Panda robot and
Meta Aria glasses. A calibrated RGB-D camera mounted on the robot captures
12 reference views of each tabletop scene at approximately $30^{\circ}$
intervals, ensuring sufficient overlap.
The headset records egocentric query images.
Multiple ArUco markers (Marker board) are placed in the workspace to provide
ground-truth headset poses. With known marker positions, extrinsic calibration can then be determined by minimizing the reprojection error over all markers \cite{7989443}.
Under the validation with robot kinematics, our setup
achieves sub-$1$mm
translational and sub-$0.1^{\circ}$ rotational errors. Markers have
also been widely adopted as external ground-truth in previous
robotics and vision evaluations~\cite{7989443,wang2026seeing}. For evaluation, we only consider frames in which a valid marker detection is available. To prevent the markers from contributing to
the results, features detected within or near the
marker regions are filtered out.
As
shown in Tab~\ref{tab:dataset_scenes} and Fig. \ref{data}, our
dataset contains 3 categories of tabletop scenes, each category contains 10 sequences. To improve generalization, the object number, position, type, height, illumination, and
scene background are varied across sequences. These sequences provide
complementary geometric conditions for evaluating different cross-device
alignment methods. Each sequence contains robot RGB-D reference images with calibrated camera
poses (12 images), egocentric headset RGB images ($>$ 40 frames), camera intrinsics, and ground-truth cross-device poses.

\section{Experiments}

\noindent\textbf{Evaluation Metrics}: We evaluate the estimated cross-device transformation using two pose-level
metrics, \textbf{Absolute Translation Error (ATE)} and \textbf{Absolute Rotation Error (ARE}). ATE measures the
translational error between the estimated cross-device transformation $\widehat{{}^{H}\mathbf{T}_{R}}$ and the ground-truth cross-device transformation ${}^{H}\mathbf{T}_{R}^{*}$, while ARE measures the rotational error. ATE and ARE quantify transformation accuracy but do not directly indicate gaze-based placement. We therefore introduce \textbf{Gaze--Surface Intersection Error (GSIE)}.
Given a gaze point ${}^{H}\mathbf{u}$ in the headset image, a 3D gaze ray
is constructed using the intrinsics ${}^{H}\mathbf{K}$ and transformed
into the robot frame. To obtain the 3D gaze target, we define
$\pi^{-1}_{\mathcal M^R}$ as the operator that returns the
3D point in the reconstructed scene $\mathcal M^R$ with the minimum
distance to the transformed viewing ray. Using the estimated pose $\widehat{{}^{H}\mathbf{T}_{R}}$ and the
ground-truth pose ${}^{H}\mathbf{T}_{R}^{*}$, we obtain two 3D gaze target
points in $\mathcal{M}^R$. GSIE is then defined as:
\vspace{-0pt}
\begin{equation}
e
=
\left\|
\pi^{-1}_{\mathcal{M}^R}
\left(
{}^{H}\mathbf{u},
{}^{H}\mathbf{K},
\widehat{{}^{H}\mathbf{T}_{R}}
\right)
-
\pi^{-1}_{\mathcal{M}^R}
\left(
{}^{H}\mathbf{u},
{}^{H}\mathbf{K},
{}^{H}\mathbf{T}_{R}^{*}
\right)
\right\|_2
\end{equation}  
GSIE therefore directly measures the spatial displacement of the gaze target
caused by cross-device alignment error. For the reference-selection experiment, we report\textbf{ Reprojection Error (RE)} to evaluate the geometric compatibility
of the selected reference with the query image. Specifically, after feature
matching and PnP estimation, the 3D points associated with the PnP inlier
correspondences are projected into the query image using the ground-truth
relative pose. \textbf{Reference Selection Accuracy (RSAcc.)} denotes the percentage
of headset query images for which the selected robot reference matches the
ground-truth reference. The ground-truth reference is defined as the robot view with the lower reprojection error to the query.

\noindent\textbf{Experimental Setup}: All experiments are conducted on a workstation equipped with a 12th Gen
Intel Core i9-12900K CPU and an NVIDIA RTX 3090 GPU. Object detection and instance segmentation are performed using YOLO26 \cite{jocher2026ultralyticsyolo26unifiedrealtime}. 
We selected the angular-bin width from $\{30^\circ,45^\circ,60^\circ,90^\circ\}$ using mean RSAcc. on a small validation split. The resulting $45^\circ$ bin width was fixed for all test scenes.
We also benchmark the methods on three sequences from the TUM RGB-D
benchmark~\cite{sturm2012evaluating}, namely \textit{fr2/desk},
\textit{fr3/long}, and \textit{fr2/dishes}. These sequences provide more
cluttered and less controlled indoor scenes than our tabletop setup and are
used to evaluate robustness and generalizability
beyond our own acquisition environment. Following our acquisition protocol,
images are sampled at approximately $30^{\circ}$ viewpoint intervals as robot
reference views, while intermediate frames are used as headset query images.
To better reproduce the asymmetric cross-device setting, the robot reference
images are cropped to retain mainly the tabletop workspace, whereas the query
images preserve a wider scene context. The camera intrinsics are adjusted
accordingly after cropping. This setup also introduces different effective
fields of view and camera parameters between the reference and query images,
consistent with the robot--headset setting.

\begin{table}[t]
    \centering
    \vspace{7pt}
    \refstepcounter{table}
    \label{multi-Alignment}

    \begin{minipage}{\columnwidth}
        \footnotesize
        \centering
        \textbf{Table~\thetable.}
        Benchmark of cross-device alignment approaches (RMSE).
    \end{minipage}
    \par\vspace{0.8ex}

    \footnotesize
    \setlength{\tabcolsep}{6pt}
    \renewcommand{\arraystretch}{1.08}

    \begin{adjustbox}{max width=\columnwidth}
        \begin{tabular}{@{}clccc@{}}
            \toprule
            \textbf{Scene}
            & \textbf{Approach}
            & \shortstack{\textbf{ATE}\\(mm)}
            & \shortstack{\textbf{ARE}\\($^\circ$)}
            & \shortstack{\textbf{GSIE}\\(mm)} \\
            \midrule

            \multirow{4}{*}{$S_A$}
            & PnP-LoMa
            & 33.3 & 1.6 & 23.9 \\
            & PnP-LightGlue
            & 32.8 & 1.5 & 23.7 \\
            & PnP+Line
            & 32.4 & 1.7 & 22.4 \\
            & Refinement via Ellipsoid
            & \textbf{29.6} & \textbf{1.3} & \textbf{22.1} \\
            \midrule

            \multirow{4}{*}{$S_B$}
            & PnP-LoMa
            & 27.4 & \textbf{1.3} & 18.3 \\
            & PnP-LightGlue
            & 28.4 & 1.4 & 18.8 \\
            & PnP+Line
            & 44.3 & 3.2 & 22.2 \\
            & Refinement via Ellipsoid
            & \textbf{21.8} & 1.6 & \textbf{17.3} \\
            \midrule

            \multirow{4}{*}{$S_C$}
            & PnP-LoMa
            & 33.0 & 2.3 & 12.7 \\
            & PnP-LightGlue
            & 32.5 & 2.2 & 12.9 \\
            & PnP+Line
            & \textbf{25.2} & \textbf{1.7} & 11.5 \\
            & Refinement via Ellipsoid
            & 28.6 & 2.0 & \textbf{9.9} \\
            \midrule

            \multirow{4}{*}{\textit{fr2/desk}}
            & PnP-LoMa
            & 40.0 & 2.9 & 150.8 \\
            & PnP-LightGlue
            & 27.8 & \textbf{2.4} & 149.9 \\
            & PnP+Line
            & 33.2 & \textbf{2.4} & 144.1 \\
            & Refinement via Ellipsoid
            & \textbf{24.1} & 2.5 & \textbf{142.7} \\
            \midrule

            \multirow{4}{*}{\textit{fr3/long}}
            & PnP-LoMa
            & \textbf{34.3} & \textbf{1.0} & 31.9 \\
            & PnP-LightGlue
            & 40.0 & 1.3 & 32.4 \\
            & PnP+Line
            & 74.9 & 3.4 & 35.5 \\
            & Refinement via Ellipsoid
            & 36.4 & 1.1 & \textbf{27.4} \\
            \midrule

            \multirow{4}{*}{\textit{fr2/dishes}}
            & PnP-LoMa
            & \textbf{42.7} & \textbf{2.2} & 23.7 \\
            & PnP-LightGlue
            & 47.1 & 2.5 & 22.7 \\
            & PnP+Line
            & 52.0 & 2.7 & 21.6 \\
            & Refinement via Ellipsoid
            & 45.3 & 2.3 & \textbf{20.8} \\
            \bottomrule
        \end{tabular}
    \end{adjustbox}

    \par\vspace{0.5ex}
    \begin{minipage}{\columnwidth}
        \scriptsize
        \raggedright
        Lower values are better.
        Best results for each metric within each scene are
        bold, including ties.
    \end{minipage}
    \vspace{-25pt}
\end{table}

\begin{table*}[t]
    \centering
    \vspace{7pt}
    \refstepcounter{table}
    \label{multi-perspective}

    \begin{minipage}{\textwidth}
        \footnotesize
        \centering
        \textbf{Table~\thetable.}
        Comparison of reference selection methods.
    \end{minipage}
    \par\vspace{0.8ex}

    \footnotesize
    \setlength{\tabcolsep}{4pt}
    \renewcommand{\arraystretch}{1.08}

    \begin{adjustbox}{max width=\textwidth}
        \begin{tabular}{@{}lccccccccccccc@{}}
            \toprule
            \multirow{2}{*}{\textbf{Scene}}
            & \textbf{GT ref.}
            & \multicolumn{3}{c}{\textbf{HF-Net}~\cite{sarlin2019coarse}}
            & \multicolumn{3}{c}{\textbf{MegaLoc}~\cite{Berton_2025_MegaLoc}}
            & \multicolumn{3}{c}{\textbf{Exhaustive Matching}}
            & \multicolumn{3}{c}{\textbf{Ours}} \\
            \cmidrule(lr){2-2}
            \cmidrule(lr){3-5}
            \cmidrule(lr){6-8}
            \cmidrule(lr){9-11}
            \cmidrule(l){12-14}

            & RE
            & RSAcc. & RE & Runtime
            & RSAcc. & RE & Runtime
            & RSAcc. & RE & Runtime
            & RSAcc. & RE & Runtime \\
            \midrule

            $S_A$
            & $\mathbf{4.4 \pm 1.7}$
            & 0.13 & $92.7 \pm 32.5$ & 4867
            & 0.87 & $26.9 \pm 19.4$ & 436
            & 0.74 & $42.3 \pm 23.6$ & 361
            & \textbf{1.00}
            & $\mathbf{4.4 \pm 1.7}$
            & \textbf{13.97} \\

            $S_B$
            & $\mathbf{5.3 \pm 1.9}$
            & 0.07 & $104.2 \pm 43.1$ & 4612
            & 0.28 & $21.3 \pm 9.6$ & 412
            & 0.19 & $39.7 \pm 35.7$ & 327
            & \textbf{0.98}
            & $6.5 \pm 2.1$
            & \textbf{14.02} \\

            $S_C$
            & $\mathbf{3.9 \pm 1.4}$
            & 0.21 & $86.4 \pm 28.6$ & 5217
            & 0.69 & $18.7 \pm 6.1$ & 459
            & 0.62 & $28.3 \pm 13.9$ & 374
            & \textbf{1.00}
            & $\mathbf{3.9 \pm 1.4}$
            & \textbf{13.98} \\
            \midrule

            \textit{fr2/desk}
            & $\mathbf{18.4 \pm 2.3}$
            & 0.17 & $124.3 \pm 29.8$ & 5134
            & 0.91 & $26.9 \pm 19.4$ & 439
            & 0.77 & $44.6 \pm 42.4$ & 394
            & \textbf{0.97}
            & $20.8 \pm 3.5$
            & \textbf{15.06} \\

            \textit{fr3/long}
            & $\mathbf{2.2 \pm 0.3}$
            & 0.19 & $4.6 \pm 1.4$ & 5269
            & 0.84 & $4.2 \pm 0.7$ & 422
            & 0.89 & $3.9 \pm 0.9$ & 387
            & \textbf{1.00}
            & $\mathbf{2.2 \pm 0.3}$
            & \textbf{15.08} \\

            \textit{fr2/dishes}
            & $\mathbf{8.9 \pm 2.7}$
            & 0.04 & $1010.1 \pm 475.3$ & 4238
            & 0.19 & $865.3 \pm 382.5$ & 447
            & 0.23 & $103.0 \pm 48.6$ & 392
            & \textbf{0.95}
            & $9.2 \pm 2.9$
            & \textbf{12.93} \\
            \bottomrule
        \end{tabular}
    \end{adjustbox}

    \par\vspace{0.5ex}
    \begin{minipage}{\textwidth}
        \scriptsize
        \raggedright
        ``GT ref.'' denotes the reference view with the lower
        reprojection error.
        RSAcc.: reference selection accuracy, reported as a fraction.
        RE is reported in pixels and runtime in milliseconds.
        All methods use the same feature matcher
        \cite{lindenberger2023lightglue} and PnP solver after
        reference selection.
        All evaluated queries produced valid PnP estimates.
        Our runtime includes YOLO26
        \cite{jocher2026ultralyticsyolo26unifiedrealtime}
        inference and graph matching.
        HF-Net and MegaLoc runtimes include descriptor extraction
        and retrieval.
        Exhaustive matching runtime includes exhaustive matching
        and PnP selection.
    \end{minipage}
    \vspace{-5pt}
\end{table*}

\begin{figure*}
    \centering
    \includegraphics[width=0.95\textwidth]{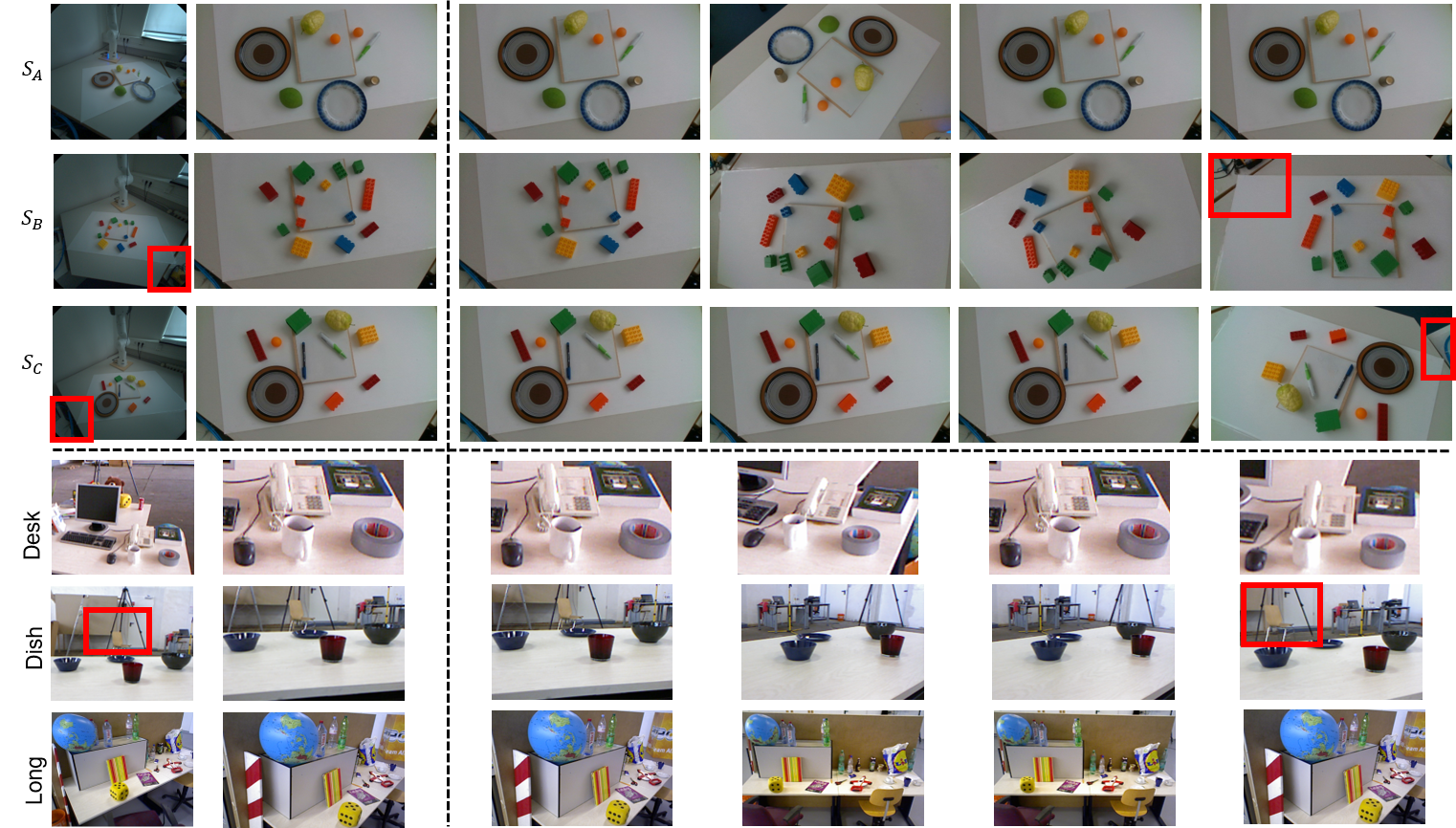}
    \vspace{-5pt}
    \caption{Qualitative comparison of reference selection.
    The red boxes indicate feature-rich background regions,
    and exhaustive matching tends to select references that
    match these regions.
    Our method maintains robust object association under
    weak object textures.}
    \label{feature}
    \vspace{-15pt}
\end{figure*}

\subsection{Benchmark of Cross-Device Alignment}
Tab~\ref{multi-Alignment} evaluates four cross-device alignment pipelines
that we develop or adapt specifically for gaze-based manipulation, covering
point-based matching, line-assisted optimization, and object-level geometric
refinement. Such a unified benchmark
is scarce, since existing methods are typically designed for localization rather than robot--headset alignment with a downstream gaze-placement
objective. To isolate the evaluation from additional errors of gaze point estimation, GSIE in this
benchmark is computed over all valid points in each headset query image rather
than using an estimated measured gaze point. A point is considered valid
if has available depth in the
reconstructed scene $\mathcal{M}^{R}$. 

The two PnP pipelines provide a relatively stable baseline. Their performance is mainly determined by the quality
and spatial distribution of the feature correspondences. For ATE and ARE, LoMa \cite{nordstrom2026loma} performs
better in several challenging scenes, such as $S_B$, \textit{fr2/dishes}, and \textit{fr3/long},
indicating improved robustness to larger viewpoint changes, whereas
SuperPoint+LightGlue (SP+LG) \cite{lindenberger2023lightglue} is superior in others. Overall, neither matcher consistently dominates, suggesting
that feature-based PnP remains sensitive to scene texture and viewpoint
configuration. Considering the lower runtime of
SP+LG (see Sec. \ref{runtime}), we use it as the default PnP pipeline.

The benchmark also reveals that additional geometric constraints for PnP algorithm
does not improve pose estimation in every scene. PnP+Line
performs well when reliable point and line features coexist, as in $S_C$,
but degrades considerably in $S_B$, \textit{fr2/dishes}, and \textit{fr3/long}. Although $S_B$
contains many cuboid objects and therefore rich line structures, object
boundaries often coincide with depth discontinuities, where RGB-D
measurements are noisy, missing, or affected by occlusion. Consequently,
the reconstructed 3D lines can introduce wrong constraints, particularly
in scenes with rich line structures but weak textures ($S_B$ and  \textit{fr2/dishes}). In contrast, line-based
VO systems such as AirVO~\cite{xu2023airvo} recover 3D line geometry through
stereo or multi-view observations and jointly optimize it across frames,
rather than relying directly on noisy depth measurements at object boundaries.
At the other extreme, \textit{fr3/long} contains relatively few reliable line
features, limiting the benefit of line features.


More importantly, the benchmark reveals a clear mismatch between pose-space
metric and task-oriented metric. Ellipsoid refinement achieves the lowest
GSIE in every scene, although it does not consistently achieve the lowest
ATE or ARE. In $S_C$, for example, PnP+Line obtains the best ATE and ARE
($25.2$\,mm and $1.7^\circ$), yet ellipsoid refinement achieves a lower GSIE
($9.9$\,mm versus $11.5$\,mm) despite worse pose errors. Similarly, in
\textit{fr3/long} and \textit{fr2/dishes}, PnP-LoMa achieves the best ATE and
ARE, while ellipsoid refinement achieves the lowest GSIE. By aligning projected
3D quadrics with observed 2D object geometry, the refinement constrains the
cross-device pose using scene-level object structure, which better preserves
the gaze-to-scene mapping required for accurate placement.
Across all method--scene combinations, GSIE exhibits only weak correlations
with conventional pose metrics (ATE: Pearson $r=-0.12$, Spearman $\rho=0.31$;
ARE: Pearson $r=0.36$, Spearman $\rho=0.21$, all $p>0.05$), with none reaching statistical significance. This discrepancy arises because ATE and ARE capture only pose-error
magnitudes, not their directions relative to the gaze ray and scene geometry. The placement position is jointly determined by translation,
rotation, viewing direction, and target depth; translational and rotational
errors may therefore compensate for or amplify each other.
GSIE captures this coupling directly in task space and is therefore necessary
for evaluating the actual quality of gaze-based placement.

\begin{table*}[t]
    \centering
    \vspace{7pt}
    \refstepcounter{table}
    \label{multi-interaction}

    \begin{minipage}{\textwidth}
        \footnotesize
        \centering
        \textbf{Table~\thetable.}
        Comparison of gaze-based interaction in real-world scenes.
    \end{minipage}
    \par\vspace{0.8ex}

    \footnotesize
    \setlength{\tabcolsep}{4pt}
    \renewcommand{\arraystretch}{1.08}

    \begin{adjustbox}{max width=\textwidth}
        \begin{tabular}{@{}llccccccccc@{}}
            \toprule
            \multirow{2}{*}{\textbf{Approach}}
            & \multirow{2}{*}{\textbf{Paradigm}}
            & \multirow{2}{*}{\shortstack{\textbf{Arbitrary}\\\textbf{position?}}}
            & \multirow{2}{*}{\shortstack{\textbf{Arbitrary}\\\textbf{object?}}}
            & \multirow{2}{*}{\textbf{Occlusion?}}
            & \multicolumn{3}{c}{\textbf{Selection Accuracy}}
            & \multicolumn{3}{c}{\textbf{Placement Success Rate}} \\
            \cmidrule(lr){6-8}
            \cmidrule(l){9-11}

            & & & &
            & $S_A$ & $S_B$ & $S_C$
            & $S_A$ & $S_B$ & $S_C$ \\
            \midrule

            GOReloc~\cite{wang2024goreloc}
            & Graph + PnP
            & \textcolor{green}{\ding{51}}
            & \textcolor{green}{\ding{51}}
            & \textcolor{green}{\ding{51}}
            & 0.74 & 0.53 & 0.69
            & 0.34 & 0.38 & 0.36 \\

            AR Marker~\cite{9889538}
            & ArUco alignment
            & \textcolor{green}{\ding{51}}
            & \textcolor{green}{\ding{51}}
            & \textcolor{red}{\ding{55}}
            & 0.89 (1.00) & 0.85 (1.00) & 0.91 (1.00)
            & 0.88 (1.00) & 0.83 (1.00) & 0.90 (1.00) \\

            FAM-HRI~\cite{lai2026fam}
            & Local feature matching
            & \textcolor{gray}{%
                \ding{51}%
                \rotatebox[origin=c]{-9.2}{\kern-0.60em\ding{55}}%
              }
            & \textcolor{green}{\ding{51}}
            & \textcolor{green}{\ding{51}}
            & 0.57 & 0.63 & 0.58
            & 0.28 & 0.16 & 0.31 \\

            Label-Based~\cite{shahid2025gear}
            & Object labels
            & \textcolor{red}{\ding{55}}
            & \textcolor{red}{\ding{55}}
            & \textcolor{green}{\ding{51}}
            & 0.61 & 0.23 & 0.53
            & -- & -- & -- \\

            \midrule
            \textbf{Ours}
            & Graph + refinement
            & \textcolor{green}{\ding{51}}
            & \textcolor{green}{\ding{51}}
            & \textcolor{green}{\ding{51}}
            & \textbf{0.96} & \textbf{0.91} & \textbf{0.93}
            & \textbf{0.94} & \textbf{0.91} & \textbf{0.92} \\
            \bottomrule
        \end{tabular}
    \end{adjustbox}

    \par\vspace{0.5ex}
    \begin{minipage}{\textwidth}
        \scriptsize
        \raggedright
        Accuracy and success rates are reported as fractions.
        ``--'' indicates that the metric cannot be measured.
        Parenthesized values report AR-marker results after
        excluding occlusions and missing detections.
        Bold marks the best non-parenthesized value in each
        metric column.
    \end{minipage}
    \vspace{-25pt}
\end{table*}

\subsection{Performance of Graph-based Reference Selection}
\label{Performance of Graph-based Reference Selection}

We compare our graph-based reference selection with
HF-Net~\cite{sarlin2019coarse}, MegaLoc~\cite{Berton_2025_MegaLoc}, and
exhaustive matching. Exhaustive matching
evaluates all reference images and selects the one producing the most PnP
inliers. As shown in Tab~\ref{multi-perspective}, our method achieves an RSAcc. of
0.95--1.00 across all scenes, the lowest RE, and a runtime of approximately
15 ms, demonstrating both efficient and reliable reference
selection.

The lower RSAcc. of HF-Net and MegaLoc can be attributed to the
difference between appearance similarity and viewpoint compatibility. As shown in Fig. \ref{feature}, both
methods tend to retrieve the image that is globally most similar to the query,
but the most appearance-similar image is not always captured from the most compatible viewing direction. Exhaustive matching considers every reference candidate, but selecting the
candidate with the largest number of PnP inliers does not always identify
the most compatible reference. The inlier count can be dominated by
feature-rich regions rather than the objects
relevant to cross-device alignment. This issue is particularly evident in
$S_B$ and \textit{fr2/dishes}, where the objects contain relatively
few detectable features. In these scenes, a feature-rich region may produce more inliers while still being
geometrically unsuitable for alignment. In contrast, our semantic-spatial graph emphasizes the
object configuration and viewing-direction compatibility, enabling it to
select geometrically useful references without exhaustively matching.

\subsection{Baseline Comparison}
We evaluate gaze-based placement in the real-world scenes of our dataset.
As shown in Fig. \ref{data} (right), the user wears Meta Aria glasses and gazes at the target location, while specifying an object to be
moved. ``Selection Acc." measures whether the intended object is correctly
identified, while for ``Placement Success Rate", a trial is considered successful when the midpoint of the grasping points lies within the 5cm radius target region. An operator recorded 100 independent gaze–speech commands in each scene. The same recorded queries, initial object configurations, and target locations were evaluated by all five methods. All object and target
positions are selected within the verified reachable workspace of the robot
to avoid failures caused by kinematic limitations. The target regions are masked out for all methods.

As shown in Tab \ref{multi-interaction}, GOReloc~\cite{wang2024goreloc} is a graph-based cross-view alignment method that
associates objects through graph matching. FAM-HRI~\cite{lai2026fam} relies entirely on local feature correspondences. It associates each query object
with the reference object having the largest number of local feature
correspondences within its object region. AR Marker \cite{9889538} and Ours first estimate the relative pose and associate detected
and projected object bounding boxes using the IoU and Hungarian Algorithm.
For GOReloc, its graph compares geometric relations between 2D detections and a 3D
object map, which are affected by perspective projection. Moreover, its pose
initialization relies on correspondences between 2D object centers and 3D
object centroids, resulting in coarse geometric constraints and relatively
low physical placement success rates. The AR-marker baseline provides ground-truth alignment
accuracy whenever the marker board is successfully detected. However,
its failures are dominated by occlusion and missed detections, making the
method sensitive to marker visibility and restricting the usable motion range.
For FAM-HRI, under large cross-view angles, especially for weakly textured objects, there are few
reliable object-level matches remain; relaxing the matching threshold improves
coverage but also increases incorrect associations. For placement, local
features around the gaze target are matched to the reference view and the
centroid is used as the target position, which becomes unstable when the matches
are sparse or incorrect. Label-based interaction~\cite{shahid2025gear} is
limited to objects with distinguishable labels and cannot resolve
repeated labels or arbitrary placement. In contrast, our method
achieves $0.91-0.96$ selection accuracy and $0.91-0.94$ placement success
rate while supporting arbitrary placement positions without
requiring visible markers. The selection errors are mainly caused by missed object detections.

\begin{table}[tb]
    \centering
    \vspace{7pt}
    \refstepcounter{table}
    \label{multi-runtime}

    \begin{minipage}{\columnwidth}
        \footnotesize
        \centering
        \textbf{Table~\thetable.}
        Runtime analysis.
    \end{minipage}
    \par\vspace{0.8ex}

    \footnotesize
    \setlength{\tabcolsep}{6pt}
    \renewcommand{\arraystretch}{1.08}

    \begin{adjustbox}{max width=\columnwidth}
        \begin{tabular}{@{}lcc@{}}
            \toprule
            \textbf{Component}
            & \shortstack{\textbf{Cross-device dataset}\\(ms)}
            & \shortstack{\textbf{TUM-RGBD}\\(ms)} \\
            \midrule

            Detection \& Segmentation
            & 11.03 & 11.57 \\

            Graph-based Selection
            & 2.98 & 3.06 \\

            Feature Matching
            & 19.53 & 19.14 \\

            PnP
            & 1.07 & 1.02 \\

            PnP+Line
            & 3.11 & 3.35 \\

            Ellipsoid Refinement
            & 6.13 & 6.25 \\
            \bottomrule
        \end{tabular}
    \end{adjustbox}

    \par\vspace{-25pt}
    
\end{table}

\subsection{Runtime Analysis}
\label{runtime}

Tab~\ref{multi-runtime} reports the runtime of each component.
Graph-based reference selection introduces only about $3$\,ms overhead,
while SuperPoint+LightGlue \cite{lindenberger2023lightglue} feature matching accounts for most of the
computation. The complete pipelines
all operate at approx. $24$ FPS, demonstrating real-time performance. In comparison, the frame rate for 
egocentric RGB camera on Aria glasses is $20$ FPS.   
We also evaluated more recent wide-baseline matchers such as LoMa \cite{nordstrom2026loma}.
Although such methods can improve robustness under challenging viewpoint
changes, our implementation required approximately $126$\,ms per image pair
even with ONNX acceleration.

 

\subsection{Ablation Study}

\begin{table}[tb]
    \centering
    \vspace{7pt}
    \refstepcounter{table}
    \label{multi-ablation}

    \begin{minipage}{\columnwidth}
        \footnotesize
        \centering
        \textbf{Table~\thetable.}
        Alignment ablation averaged over $S_A$, $S_B$, and $S_C$.
    \end{minipage}
    \par\vspace{0.8ex}

    \footnotesize
    \setlength{\tabcolsep}{8pt}
    \renewcommand{\arraystretch}{1.08}

    \begin{adjustbox}{max width=\columnwidth}
        \begin{tabular}{@{}lcccc@{}}
            \toprule
            \textbf{Approach}
            & \shortstack{\textbf{ATE}\\(mm)}
            & \shortstack{\textbf{ARE}\\($^\circ$)}
            & \shortstack{\textbf{GSIE}\\(mm)}
            & \shortstack{\textbf{GSIE change}\\(\%)} \\
            \midrule

            \multicolumn{5}{c}{\textit{Enclosing Ellipsoid}} \\
            \midrule

            PnP-baseline
            & 31.2 & 1.7 & 18.5 & 0.0 \\

            PCA-based fitting~\cite{voom}
            & 29.7 & 1.7 & 17.8 & $-3.8$ \\

            Least-Shell-Distance~\cite{1290055}
            & 28.9 & 1.7 & 17.2 & $-7.0$ \\

            MVEE
            & \textbf{26.7}
            & \textbf{1.6}
            & \textbf{16.4}
            & $\mathbf{-11.4}$ \\
            \midrule

            \multicolumn{5}{c}{%
                \textit{Feature Extraction \& Matching}} \\
            \midrule

            ORB~\cite{mur2015orb}
            & \textbf{30.6}
            & 3.1
            & 23.1
            & $+24.9$ \\

            SIFT+LG
            & 33.7
            & 1.8
            & 20.1
            & $+8.6$ \\

            SP+LG
            & 31.2
            & \textbf{1.7}
            & \textbf{18.5}
            & \textbf{0.0} \\
            \bottomrule
        \end{tabular}
    \end{adjustbox}

    \par\vspace{0.5ex}
    \begin{minipage}{\columnwidth}
        \scriptsize
        \raggedright
        GSIE change is computed as
        $100(g-g_{\mathrm{base}})/g_{\mathrm{base}}$,
        using PnP-baseline for the ellipsoid block and
        SP+LG for the feature extraction and matching block.
        Percentage changes are calculated from the displayed
        mean GSIE values; negative values indicate improvement.
        Best results within each block are bold.
    \end{minipage}
    \vspace{-15pt}
\end{table}



\begin{table}[tb]
    \centering
    \vspace{7pt}
    \refstepcounter{table}
    \label{ablationgraph}

    \begin{minipage}{\columnwidth}
        \footnotesize
        \raggedright
        \textbf{Table~\thetable.}
        Averaged reference selection results with descriptor ablations.
    \end{minipage}
    \par\vspace{0.8ex}

    \footnotesize
    \setlength{\tabcolsep}{9pt}
    \renewcommand{\arraystretch}{1.08}

    \begin{adjustbox}{max width=\columnwidth}
        \begin{tabular}{@{}lcccc@{}}
            \toprule
            \multirow{2}{*}{\textbf{Descriptor}}
            & \multicolumn{2}{c}{\textbf{Cross-device dataset}}
            & \multicolumn{2}{c}{\textbf{TUM-RGBD}} \\
            \cmidrule(lr){2-3}
            \cmidrule(l){4-5}

            & RSAcc. & RE (px)
            & RSAcc. & RE (px) \\
            \midrule

            w/o Semantics
            & 0.84
            & $29.3 \pm 11.9$
            & 0.73
            & $89.6 \pm 64.8$ \\

            w/o Angle
            & 0.13
            & $128.7 \pm 94.1$
            & 0.07
            & $794.3 \pm 412.2$ \\

            With Distance Bins
            & 0.38
            & $78.1 \pm 53.6$
            & 0.11
            & $687.1 \pm 572.0$ \\

            \midrule
            \textbf{Full Descriptor}
            & \textbf{0.99}
            & $\mathbf{4.8 \pm 2.4}$
            & \textbf{0.97}
            & $\mathbf{12.3 \pm 6.1}$ \\
            \bottomrule
        \end{tabular}
    \end{adjustbox}

    \par\vspace{0.5ex}
    \begin{minipage}{\columnwidth}
        \scriptsize
        \raggedright
        RSAcc.: reference selection accuracy, reported as a fraction.
        RE: reprojection error.
        Higher RSAcc.\ and lower RE are better.
        Best results in each metric column are bold.
    \end{minipage}
    \vspace{-25pt}
\end{table}

Tab.~\ref{multi-ablation} reports the averaged alignment
results over $S_A-S_C$. All ellipsoid refinements improve the PnP baseline, with MVEE achieving the
lowest error in all metrics. In particular, MVEE reduces ATE from
$31.2$mm to $26.7$mm and GSIE from $18.5$mm to $16.4$mm.
ORB slightly improves ATE but substantially degrades ARE and
GSIE, while SIFT+LightGlue underperforms the default feature
pipeline.

Tab.~\ref{ablationgraph} evaluates the graph descriptor. In ``w/o Semantic'', class conditioning is removed; in
``w/o Angle'', all angular bins are collapsed; and in
``Distance Bins'', angular relations are replaced with
normalized pairwise-distance bins. All other pipeline
components remain unchanged.
Removing either semantic or angular information substantially
degrades reference selection. Replacing angular relations with
binned image distances
remains considerably worse than the full descriptor. These results show
that semantic conditioning and coarse angular relations provide
complementary cues for cross-view reference selection.

\section{Conclusion}
In this work, we present a markerless framework for gaze-based object placement at
arbitrary positions, together with a dedicated cross-device dataset.
To address sparse reference views and large robot--headset viewpoint
differences, we propose Graph-based Reference Selection, which combines
semantic structural reasoning with feature-based geometric alignment.
We also benchmark point-, line-, and
object-based refinement pipelines. We further show that ATE and ARE do not fully characterize
the downstream placement error and introduce GSIE to directly measure the
deviation of the gaze--surface intersection. 
Future work will investigate
alignment objectives that explicitly incorporate GSIE into
end-to-end optimization for gaze-based placement.
\section*{Acknowledgments}
\noindent AI Usage Statement. Codex assisted with code development. ChatGPT assisted with manuscript drafting and editing. 
\bibliographystyle{IEEEtran}
\bibliography{IEEEexample}

\end{document}